\pdfoutput=1

\documentclass[11pt]{article}

\usepackage[]{naacl2021}

\usepackage{times}
\usepackage{latexsym}
\usepackage{amsmath, amssymb, amsfonts, bm}
\usepackage{graphicx}
\usepackage{arydshln}
\usepackage{multirow}
\usepackage{subcaption}

\usepackage[T1]{fontenc}

\usepackage[utf8]{inputenc}

\usepackage{microtype}

\title{Adversarial Learning for Zero-Shot Stance Detection on Social Media}

\author{
Emily Allaway\thanks{$^*$ Denotes equal contribution.} \And Malavika Srikanth$^*$\\
    Department of Computer Science, Columbia University, New York, NY\\
  \texttt{\{eallaway,kathy\}@cs.columbia.edu, \tt ms5908@columbia.edu} \\
  \And Kathleen McKeown\\
  \\}

\begin{document}
\maketitle
\begin{abstract}
Stance detection on social media can help to identify and understand slanted news or commentary in
everyday life. In this work, we propose a new model for zero-shot stance detection on Twitter that uses adversarial learning to generalize across topics. Our model achieves state-of-the-art performance on a number of unseen test topics with minimal computational costs. In addition, we extend zero-shot stance detection to new topics, highlighting future directions for zero-shot transfer. 
\end{abstract}

\section{Introduction}
Stance detection, the problem of automatically identifying positions or opinions in text, is becoming increasingly important for social media (e.g., Twitter), as more and more people turn to it for their news. Zero-shot stance detection, in particular, is crucial, since gathering training data for all topics is not feasible. 
While there has been increasing work on zero-shot stance detection in other genres~\citep{Allaway2020ZeroShotSD,Vamvas2020XA}, generalization across many topics in social media remains an open challenge. 

In this work, we propose a new model for stance detection that uses adversarial learning to generalize to unseen topics on Twitter. Our model achieves state-of-the-art zero-shot performance on the majority of topics in the standard dataset for English stance detection on Twitter ~\citep{Mohammad2016SemEval2016T6} and also provides benchmark results on two
new topics in this dataset. 

Most prior work on English social media stance detection uses the SemEval2016 Task 6 (SemT6) dataset~\citep{Mohammad2016SemEval2016T6} which consists of six topics. While early work trained using five topics and evaluated on the sixth (e.g.,~\citet{Augenstein2016StanceDW,Zarrella2016MITREAS,Wei2016pkudblabAS}), they used only one topic, `Donald Trump' (DT), for evaluation and did not experiment with others. Furthermore, recent 
work on SemT6 has focused on \textit{cross-target} stance detection~\citep{Xu2018CrossTargetSC,Wei2019ModelingTT,Zhang2020EnhancingCS}: training on \textit{one} topic and evaluating on \textit{one} different unseeen topic that has a \textit{known relationship} with the training topic 
(e.g., ``legalization of abortion'' to ``feminist movement''). These models are typically evaluated on four different test topics (each with a different training topic). 

In contrast, our work is a hybrid of these two settings: we train on five topics and evaluate on one other, but unlike prior work we do not assume a relationship between training and test topics and so we use \textit{each topic} in turn as the test topic. This illustrates the robustness of our model across topics and additionally allows zero-shot evaluation of SemT6 on two new topics that were previously ignored by cross-target models (`atheism' and `climate change is a real concern'). 

Recently,~\citet{Allaway2020ZeroShotSD} introduced a new dataset of news article comments for zero-shot stance detection. While this dataset evaluates generalization to \textit{many new topics} when learning with many topics and only a \textit{few examples} per topic, there are no datasets for social media with this setup. Specifically, current datasets for stance detection on Twitter~\citep{Mohammad2016SemEval2016T6,Taul2017OverviewOT,Kk2017StanceDI,Tsakalidis2018NowcastingTS,Lai2020MultilingualSD} have only a \textit{few topics} but \textit{many examples} per topic. 
Therefore, zero-shot stance detection on social media is best modeled as a domain adaptation task.

To model zero-shot topic transfer as domain-adaptation, we treat each topic as a domain. Following the success of adversarial learning for domain adaptation~\citep{Zhang2017AspectaugmentedAN,Ganin2015UnsupervisedDA}, we use a discriminator (adversary) to learn topic-invariant representations that allow better generalization across topics. Although,~\citet{Wei2019ModelingTT} also proposed adversarial learning for stance detection, their model relies on knowledge transfer between topics (domains) and so is only suited to the cross-target, not zero-shot, task. In contrast, our work adopts a successful cross-target architecture into a domain adaptation model without requiring 
\textit{a priori} knowledge of
any relationship between topics. 

Our contributions in this work are: 1) we propose a new model for zero-shot stance detection on Twitter using adversarial learning that does not make assumptions about the training and test topics, and 2) we achieve state-of-the-art performance on a range of topics and provide benchmark zero-shot results for two topics not previously used in the zero-shot setting with reduced computational requirements compared to pre-trained language models. Our models are available at: \url{https://github.com/MalavikaSrikanth16/adversarial-learning-for-stance}.

\section{Methods}
\begin{figure*}
    \centering
    \vspace{-20pt}
    \includegraphics[width=0.95\textwidth]{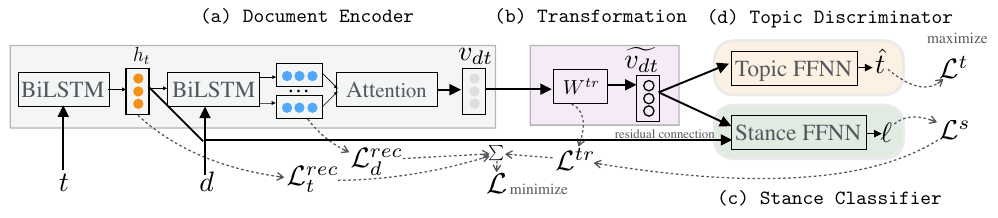}
    \caption{\textbf{TO}pic-\textbf{AD}ersarial Network (\S \ref{sec:model}). $t$ is the topic, $d$  is the document.}
    \label{fig:arch}
\end{figure*}

We propose a new model, \textbf{TO}pic-\textbf{AD}versarial Network, for zero-shot stance detection, that 
uses the domain-transfer architecture from~\citet{Zhang2017AspectaugmentedAN} coupled with a successful stance model~\citep{Augenstein2016StanceDW} with an additional topic-specific attention layer, to produce topic-invariant representations that generalize to unseen topics (see Figure~\ref{fig:arch}). 

\subsection{Overview and Definitions}
Let $D$ be a dataset of examples, each consisting of a document $d$ (a tweet), a topic $t$, and a stance label $y$. The task is to predict a label $\hat{y} \in \{$pro, con, neutral$\}$, given $d$ and $t$. 

In domain-adaptation, adversarial learning forces the model to learn domain-invariant (i.e., topic-invariant) features that can then be transferred to a new domain. To do this, a classifier and a discriminator (\textit{adversary}) are trained jointly from the same feature representation to maximize the classifier's performance while simultaneously minimizing the discriminator's. 

\subsection{Model Components}
\label{sec:model}
\textbf{(a) Topic-oriented Document Encoder}
We encode each example $x = (d, t, y)$ using bidirectional conditional encoding (BiCond) \cite{Augenstein2016StanceDW}, since computing representations conditioned on the topic 
have been shown to be crucial for zero-shot stance detection~\citep{Allaway2020ZeroShotSD}. Specifically, we first encode the topic as $h_t$ using a BiLSTM~\citep{Hochreiter1997LongSM} and then encode the text using a second BiLSTM conditioned on $h_t$. 

To compute a document-level representation $v_{dt}$, we apply scaled dot-product attention~\cite{VaswaniAttention} over the output of the text BiLSTM, using the topic representation $h_t$ as the query. This encourages the text encoder to produce representations that are indicative of stance on the topic and so would improve classification performance.

To prevent the adversary corrupting the encoder to reduce its own performance, we add a document reconstruction term ($\mathcal{L}_d^{rec}$) to our loss function, as in~\citet{Zhang2017AspectaugmentedAN}, as well as a topic reconstruction term ($\mathcal{L}_t^{rec}$), to ensure the output of neither BiLSTM is corrupted. We use a non-linear transformation over the hidden states of each BiLSTM for reconstruction. 
The reconstruction loss 
is the mean-squared error between the reconstructed vectors and the original vectors, under the same non-linearity. \\

\noindent
\textbf{(b) Topic-invariant Transformation}
To allow the adversary to produce topic-invariant representations without removing stance cues and without large adjustments to $v_{dt}$, we follow~\citet{Zhang2017AspectaugmentedAN} and apply a linear transformation  $\widetilde{v_{dt}} = W^{tr}v_{dt}$ that we regularize ($\mathcal{L}^{tr}$) to the identity $I$ .\\

\noindent
\textbf{(c) Stance Classifier}
We use a two-layer feed-forward neural network with a ReLU activation to predict stance labels $\ell \in \{-1, 0, 1\}$. Since stance is inherently dependent on a topic, and the output of the transformation layer should be topic-invariant, we add a residual connection between the topic encoder $h_t$ and the stance classifier. That is, we concatenate $h_t$ with $\widetilde{v_{dt}}$ before classification. \\

\noindent
\textbf{(d) Topic Discriminator}
Our topic discriminator is also a two-layer feed-forward neural network with ReLU and predicts the topic $t$ of the input $x$, given the output of the transformation layer $\widetilde{v_{dt}}$. 
In order to learn representations invariant to both the source and target domains, we train the discriminator using both labeled data for the source topics from $D$ and unlabeled data $D^{ul}$ for the zero-shot topic (\textit{not} from the test data), following standard practice in domain adaptation~\citep{Ganin2015UnsupervisedDA,Zhang2017AspectaugmentedAN}.

\subsection{Adversarial Training}
Our model, TOAD, is trained by combining the individual component losses. For both the stance classifier and topic-discriminator we use cross-entropy loss ($\mathcal{L}^s$ and $\mathcal{L}^t$ respectively).
Since we hypothesize that topic-invariant representations will be well suited to zero-shot
transfer, we want to minimize the discriminator's ability to predict the topic from the input. Specifically, we minimize $\mathcal{L}^s$ while maximizing $\mathcal{L}^t$, which we do using gradient reversal during backpropagation~\citep{Ganin2015UnsupervisedDA}.
Our final loss function is then
\begin{equation*}
    \mathcal{L} = \lambda_{rec} (\mathcal{L}_d^{rec} + \mathcal{L}_t^{rec}) + \lambda_{tr} \mathcal{L}^{tr} + \mathcal{L}^s - \rho \mathcal{L}^t
\end{equation*}
where $\lambda_{rec}, \lambda_{tr}$ are fixed hyperparameters. The hyperparameter $\rho$ gradually increases across epochs, following~\citet{Ganin2015UnsupervisedDA}. All loss terms except $\mathcal{L}^s$ are computed using both labeled and unlabeled data. 

\begin{table}[t]
    \centering
    \scalebox{0.8}{
    \begin{tabular}{l|rr|l}
        \hline
        \textbf{Topic} & \# Ex & \# Unlabeled & Keywords\\
        \hline
        DT & 707 & 2194 & trump, Trump\\
        HC & 984 & 1898 & hillary, clinton\\
        FM & 949 & 1951 & femini\\
        LA & 933 & 1899 & aborti\\
        CC & 564 & 1900 & climate\\ 
        A & 733 & 1900 & atheism, atheist\\
        \hline
    \end{tabular}
    }
    \caption{Data statistics for SemT6. DT: Donald Trump, HC: Hillary Clinton, FM: Feminist Movement, LA: Legalization of Abortion, CC: Climate Change is a Real Concern, A: Atheism.}
    \label{tab:datastats}
\end{table}
\begin{table*}[ht]
    \centering
    \scalebox{0.7}{
    \begin{tabular}{l|lll|lll|lll|lll|lll|lll}
        \hline
        & \multicolumn{3}{c|}{DT} & \multicolumn{3}{c|}{HC} & \multicolumn{3}{c|}{FM} & \multicolumn{3}{c|}{LA} & \multicolumn{3}{c|}{A} & \multicolumn{3}{c}{CC}\\
        & P & C & $F_{avg}$ & P & C & $F_{avg}$ & P & C & $F_{avg}$ & P & C & $F_{avg}$ & P & C & $F_{avg}$ & P & C & $F_{avg}$\\ \hline
        
        BERT & 22.3 & 57.9 & 40.1 & 36.1 & 63.2 & 49.6 & 46.6 & 37.3 & 41.9 & 36.9 & 52.8 & 44.8 & 39.6 & 70.8 & \textbf{55.2}$^{\dagger}$ & 66.3 & 8.2 &  \textbf{37.3}\\
        
        BiCond & 17.0 & 43.9 & 30.5 & 18.9 & 46.5 & 32.7 & 31.7 & 49.5 & 40.6 & 27.1 & 41.7 & 34.4 & 2.3 & 59.7 & 31.0 & 16.5 & 13.5 & 15.0\\ \hdashline
        
        CrossN & - & - & 46.1 & - & - & 41.8 & - & - & 43.1 & - & - & 44.2 & - & - & - & - & - & -\\
        
        VTN & - & - &  47.9 & - & - & 36.4 & - & - & 47.8 & - & - & 47.3 & - & - &  - & - & - & -\\
        
        SEKT & - & - & 47.7 & - & - & 42.0 & - & - & 51.3 & - & - & \textbf{53.6} & -  & - & - & - & - & -\\ \hline
        TOAD & 40.0 & 58.9 & \textbf{49.5}$^{\dagger *}$ & 35.3 & 67.1 &  \textbf{51.2} & 41.5 & 66.7 & \textbf{54.1}$^{\dagger *}$ & 30.6 & 61.7 &  46.2$^*$ & 17.7 & 74.5 & 46.1 & 45.4 & 16.5 & 30.9\\ 
        $-$ adv & 29.0 & 54.1 &  41.5 & 32.1 & 66.4 & 49.3 & 39.8 & 46.1 & 43.0 & 32.0 & 46.4 & 39.2 & 7.5 & 72.0 & 39.8 & 37.4 & 22. 0 & 29.7\\
        \hline
    \end{tabular}
    }
    \caption{Zero-shot stance $F_{avg}$ on the test sets for six topics. $^\dagger$ indicates significance ($p < 0.005$) comparing to BERT, $^*$ indicates significance ($p < 0.005$) comparing to TOAD without adversary. P is pro, C is con. Published results are used for CrossN, VTN, and SEKT; they do not report class-wise scores.}
    \label{tab:results}
\end{table*}
\begin{table}[ht]
\centering
\scalebox{0.8}{
\begin{tabular}{ll|rr}
\hline
 &  & Homogeneity & Completeness \\ \hline
\multicolumn{1}{c}{\multirow{2}{*}{DT}} & TOAD & 0.034 & 0.034 \\
\multicolumn{1}{c}{} & $-$adv & 0.102 & 0.104 \\ \hline
\multirow{2}{*}{HC} & TOAD & 0.118 & 0.120 \\
 & $-$adv & 0.135 & 0.142 \\
 \hline
\end{tabular}


}
\caption{Results of Kmeans clustering using the representations of models trained with zero-shot test topics DT and HC. Higher numbers indicates better match between the clustering and gold topic labeling.}
\label{tab:clustering}
\end{table}
\section{Experiments}
\textbf{Data}
In our experiments, we use the SemT6 dataset (see Table~\ref{tab:datastats}) used in cross-target studies~\citep{Mohammad2016SemEval2016T6}. For each topic $t \in T$, we train one model with $t$ as the zero-shot test topic. Specifically, we use all examples from each of the five topics in $\{T - t\}$ for training and validation (split $85/15$) and test on all examples for $t$. To train the topic-discriminator, we additionally use ${\sim}2k$ unlabeled tweets for the zero-shot topic $t$ from the set collected by~\citet{Augenstein2016StanceDW}. Theses tweets are from the same time period as the SemT6 dataset (${\sim}2016$) and therefore are better suited for training a discriminator than newly scraped Tweets. To select Tweets for each topic we use 1-2 keywords (see Table~\ref{tab:datastats}).\\

\noindent
\textbf{Baselines}
We compare against a BERT~\citep{Devlin2019BERTPO} baseline that encodes the document and topic jointly for classification, as in~\citet{Allaway2020ZeroShotSD} and \textbf{BiCond} -- bidirectional conditional encoding (\S \ref{sec:model}) without attention~\citep{Augenstein2016StanceDW}.
Additionally, we compare against published results from three prior models: \textbf{SEKT} -- using a knowledge graph to improve topic transfer~\citep{Zhang2020EnhancingCS}, \textbf{VTN} -- adversarial learning with a topic-oriented memory network, and \textbf{CrossN} -- BiCond with an additional topic-specific self-attention layer~\citep{Xu2018CrossTargetSC}.\\

\noindent
\textbf{Hyperparameters}
We tune the hyperparameters for our adversarial model using uniform sampling on the development set with $20$ search trials. We select the best hyperparameter setting using the average rank of the stance classifier F1 (higher is better) and topic discriminator F1 (lower is better). We remove settings where the discriminator F1 is $<0.01$, under the assumption that such low performance is the result of overly corrupt representations that will not generalize. We use pre-trained 100-dimensional GloVe vectors~\citep{pennington2014glove} in our models. 

Our implementations of BERT and BiCond are trained in the same setting as TOAD (i.e., 5 topics for train/dev, 1 topic for test). 
However, because CrossN, VTN, and SEKT are designed to learn relationships between topics, they are not suited to the zero-shot task (only the cross-target task) and therefore we report only their published cross-target results for the topic pairs (i.e., train on one, test on the other) DT $\leftrightarrow$ HC and FM $\leftrightarrow$ LA. We note that since TOAD is trained using significantly more data, our experiments evaluate not only model architectures but also the benefit of the zero-shot setting for topic-transfer.

\section{Results}
As in prior work (e.g.,~\citet{Zhang2020EnhancingCS}) we report $F_{avg}$: the average of F1 on pro and con.

Our model TOAD achieves state-of-the-art results (see Table~\ref{tab:results}) on two (DT, FM) of the four topics used in cross-target stance detection (DT: Donald Trump, HC: Hillary Clinton, FM: Feminist Movement, LA: Legalization of Abortion). These results are statistically significant ($p < 0.005$) when compared to both the BERT baseline and to TOAD without the adversary
\footnote{SEKT code is not available for computing significance.}
. In addition we provide benchmark results on two topics (A: Atheism, CC: climate change is a real concern) that have not been used previously for zero-shot evaluation.

We also observe that TOAD is statistically indistinguishable from BERT on three additional topics (HC, LA, CC) while having only $0.5\%$ as many parameters ($600k$ versus $110$mil). 
As a result of this small size, TOAD can be trained using only the CPU and, because of it's recurrent architecture, would gain less from the increased parallel computation of a GPU (compared to a transformer-based model). Therefore, TOAD has a potentially much lower environmental impact than BERT with similar (or better) performance on five out of six zero-shot topics. \\

\begin{figure}[t]
    \centering
    \begin{subfigure}[t]{0.5\textwidth}
    \includegraphics[width=\textwidth]{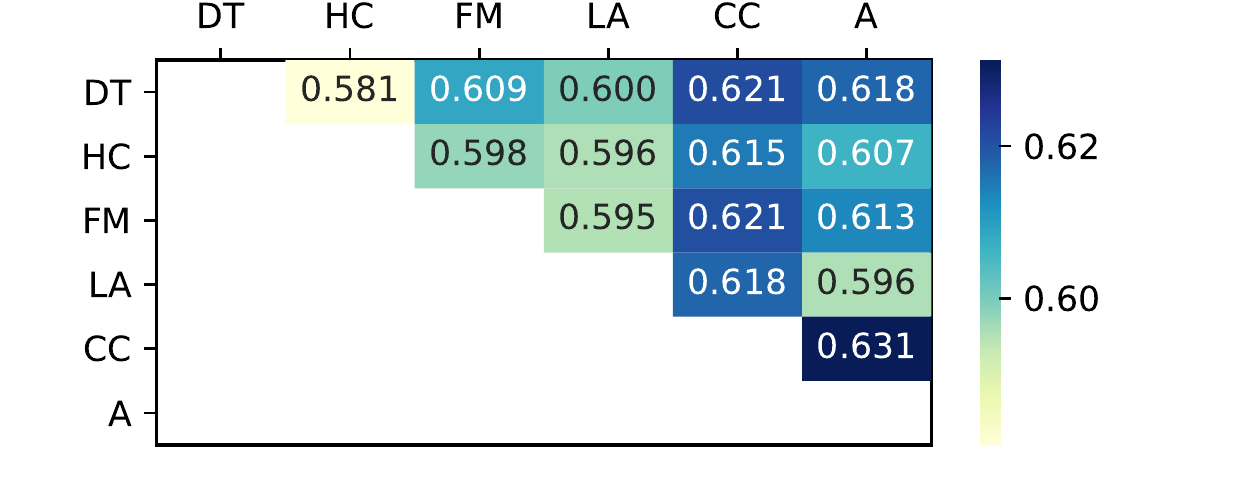}
    \caption{Using the combined vocabulary of both topics.}
    \label{subfig:bothv}
    \end{subfigure}%
    \hfill
    \begin{subfigure}[t]{0.5\textwidth}
    \includegraphics[width=\textwidth]{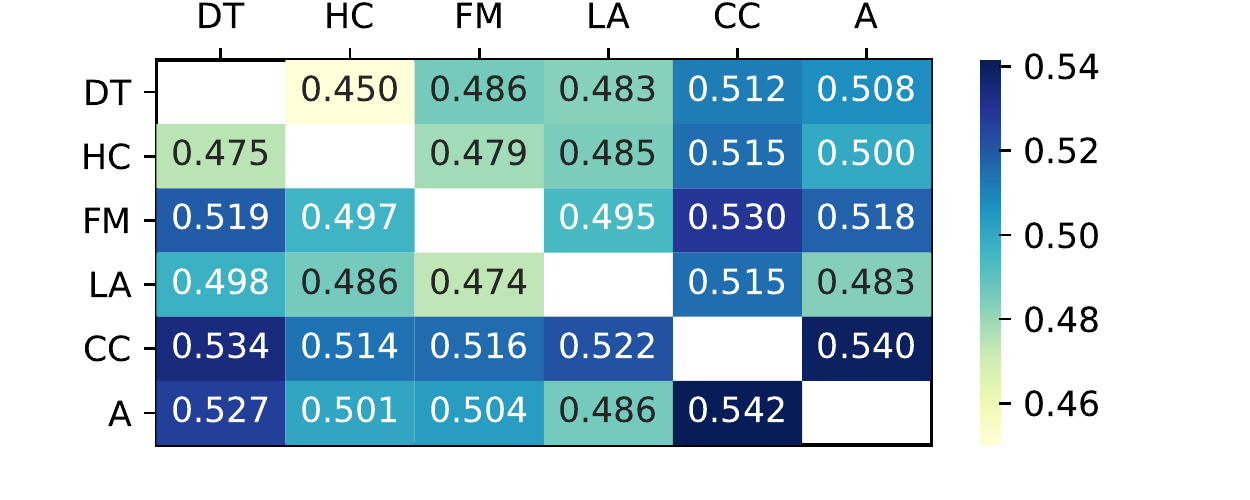}
    \caption{Using the vocabulary of the topic on the $y$-axis.}
    \label{subfig:firstv}
    \end{subfigure}
     \caption{Jensen-Shannon divergence for topic pairs.}
    \label{fig:topicdiv}
\end{figure}
\noindent
\textbf{Analysis}
Since cross-target models (e.g., SEKT) rely on assumptions about topic similarity, we first analyze the impact of topic similarity on stance performance (see Figure~\ref{fig:topicdiv}).
Specifically, we compute the Jensen-Shannon divergence~\citep{Lin1991DivergenceMB} between word distributions for pairs of topics to examine the impact of topic similarity on stance performance (see~\ref{app:topicdiv} for details). We use Jensen-Shannon divergence ($D_{JS}$) because it has been shown to successfully distinguish domains~\citep{Ruder2017LearningTS,Plank2011EffectiveMO}. 

Using the combined vocabulary of both topics in a pair (see Figure~\ref{subfig:bothv}), we observe that human notions of similarity (used to select pairs for cross-target models) may be flawed. For example, while the cross-target pair  DT $\leftrightarrow$ HC is relatively similar, for the other standard cross-target pair, FM $\leftrightarrow$ LA, FM is almost as similar to DT as to LA. Since zero-shot transfer methods use all non-test topics for training, they avoid difficulties introduced by flawed human assumptions about similarity (e.g., about the ideological similarity of FM and LA). 

We then examine, whether distributional similarity between topics does actually relate to cross-target ($T_1 \rightarrow T_2$) stance performance. Using the vocabulary for only one topic ($V_{T_1}$) per pair (see Figure~\ref{subfig:firstv}), we observe 
an inverse
relationship between similarity and relative stance performance. Specifically, relatively \textit{lower} similarity (higher divergence) often leads to relatively \textit{higher} stance performance. For example, $D_{JS}(\text{HC}||\text{DT})$ is \textit{higher} than $D_{JS}(\text{DT}|| \text{HC})$ suggesting that a model trained on HC has \textit{less} information about the word-distribution for DT than a model trained on DT has about HC. However, the cross-target stance models trained in the HC $\rightarrow$ DT setup (e.g., SEKT) actually perform relatively better than those trained in the DT $\rightarrow$ HC setup. This highlights a further problem in the cross-target setting: using similar topics may encourage models to rely on distributional patterns that do not correlate well with cross-topic stance labels.

Next, we examine
how topic-invariant the representations from TOAD actually are, and the impact of this on stance 
classification. We extract representations from our models, apply K-means clustering with $k=6$, and compare the resulting clusters to the gold topic labeling (see Table~\ref{tab:clustering}). We examine representations from models trained with either zero-shot topic DT or HC because the improvement by the adversary is statistically significant for DT but not for HC. We observe that for both topics, 
the clusters from TOAD representations are less aligned with topics.
This shows that using adversarial learning produces more topic-invariant representations than without it. 

Furthermore, we see that the difference (in both homogeneity and completeness) between TOAD with and without the adversary is larger on DT than on HC ($\Delta \approx 0.7$ and $\Delta \approx 0.02$ respectively). This suggests that the stance detection performance difference between TOAD with and without the adversary is tied to the success of the adversary at producing topic-invariant representations. That is, when the adversary is less successful, it does not provide much benefit to TOAD. 

\begin{table}[t]
    \centering
    \scalebox{0.8}{
    
    \begin{tabular}{l|rr|rr}
        \hline
        & \multicolumn{2}{c|}{DT} & \multicolumn{2}{c}{HC} \\
        & $F_{avg}$ & $\Delta$ & $F_{avg}$ & $\Delta$ \\ \hline
        TOAD & 49.5  & & 51.2 & \\ \hdashline
        $- \mathcal{L}_t^{rec}$ & 44.6 & -4.9 & 52.5 & +1.3\\
        $-$ residual topic & 39.3 & -10.2 & 43.4 & -7.8\\
        $- D^{ul}$ & 40.0 & -9.5 & 51.1 & -0.1\\
        \hline
    \end{tabular}
    }
    
    \caption{Ablation of TOAD with test sets DT and HC.}
    \label{tab:ablation}
\end{table}
Finally, we conduct an ablation on the topic-specific components of TOAD (Table~\ref{tab:ablation}). We observe that the residual topic and unlabeled data are especially important. 
Note that while the keywords used to collect unlabeled data may favor the pro class 
(e.g., \textit{aborti}), we do not observe a preference for the pro class in our models, likely due to class imbalance (e.g., 20.9\% pro DT).
Additionally, we observe that while the topic reconstruction $\mathcal{L}_t^{rec}$ is important for DT, it actually decreases the performance of the HC model. We hypothesize that this is because the adversary is less successful for HC and therefore $\mathcal{L}_t^{rec}$ only increases the noise in the stance classification loss for HC. 
Our results reaffirm the dependence of stance on the topic while also highlighting the importance of 
fully topic-invariant representations
in order to generalize.

\section{Conclusion}
We propose a new model for zero-shot stance detection on Twitter that uses adversarial learning to produce topic-invariant representations that generalize to unseen topics. Our model achieves state-of-the-art performance on a number of unseen topics with reduced computational requirements. In addition, our training procedure allows the model to generalize to new topics unrelated to the training topics and 
to provide benchmark results on two topics that have not previously been evaluated on in zero-shot settings. In future work, we plan to investigate how to extend our models to Twitter datasets in languages other than English.

\section*{Acknowledgements}
We thank the Columbia NLP group and the anonymous reviewers for their comments.
This work is supported in part by DARPA
under agreement number FA8750-18-2-0014 and by the National Science Foundation Graduate Research Fellowship under Grant No. DGE-1644869. The U.S. Government is authorized to reproduce and distribute reprints for Governmental purposes notwithstanding any copyright notation thereon. The views and conclusions contained herein are those of the authors and should not be interpreted as necessarily representing the official policies or endorsements
of DARPA, the U.S. Government, or the NSF.

\section{Ethics Statement}
We use a dataset collected and distributed for the SemEval2016 Task 6~\citep{Mohammad2016SemEval2016T6} and used extensively by the community. Data was collected from publicly available posts on Twitter using a set of manually identified hashtags (e.g., ``\#NoMoreReligions'' and ``\#Godswill'', see \url{http://saifmohammad.com/WebDocs/Stance/hashtags_all.txt} for a complete list).
 
All tweets with the hashtag at the end were collected and then post-processed to remove the actual hashtag. Thus, there is no information on the gender, ethnicity or race of the people who posted. Many of the tweets that we examined were Standard American English coupled with internet slang.

The intended use of our technology is to predict the stance of authors towards topics, where the topics are often political in nature. This technology could be useful for people in office who want to understand how their constituents feel about an issue under discussion; it may be useful to decide on new policies going forward or to react proactively to situations where people are upset about a public issue. For example, we can imagine using such a tool to determine how people feel about the safety of a vaccine or how they feel about immigration policies. If the system is incorrect in its prediction of stance, end users would not fully understand how people feel about different topics. For example, we can imagine that they may decide that there is no need to implement an education program on vaccine safety if the stance prediction tool inaccurately predicts that people feel good about vaccine safety. The benefits of understanding, with some inaccuracy, how people feel about a topic, outweigh the situation where one has no information (or only information that could be gleaned by manually reading a few examples). The technology would not be deployed, in any case, until accuracy is improved. 

We also note that since many topics are political in nature, this technology could be used nefariously to identify people to target with certain types of political ads or disinformation (based on automatically identified beliefs) or by employers to identify political opinions of employees. However, because the data does not include any user-identifying information, we ourselves are prevented from such usage and any future wrongful deployment of the technology in these settings would be a direct violation of Twitter's Terms of Service for developers\footnote{\url{https://developer.twitter.com/en/developer-terms/agreement-and-policy}}.

Given that we don't know the race of posters and we don't know whether African American Vernacular is fairly represented in the corpus, we don't know whether the tool would make fair predictions for people who speak this dialect. Further work would need to be done to create a tool that can make fair predictions regardless of race, gender or ethnicity. 

As noted in the paper, the environmental impact of training and deploying our tool is less than for all comparably performing models.

\bibliography{anthology,custom}
\bibliographystyle{acl_natbib}

\clearpage
\newpage

\appendix

\section{Appendix}
\label{sec:appendix}

\subsection{Implementation Details}
Our models are implemented using Pytorch\footnote{\url{https://pytorch.org/}}. We implement our K-means clustering with Scikit-learn\footnote{\url{https://scikit-learn.org/stable/}}. Our models are trained using one Titan Xp P8 GPU, but, as noted in the paper, they can also be trained on the CPU with a minimual increase in computation time. 

We train TOAD, TOAD without adversary, and BiCond for a maximum of $100$ epochs with early stopping on the development set, computed using $F_{avg}$. We use Adam~\citep{Kingma2015AdamAM} to optimize and for the adversarial model we decay the learning rate in relation to the discriminator strength hyperparameter $\rho$~\citep{Ganin2015UnsupervisedDA}. Specifically, until epoch 50, the learning rate is fixed at $l$ and the value of $\rho$ remains 0. If total number of epochs is $t$, for an epoch $e > 50$ we compute, $p = (e - 50)/t$.
The learning rate at epoch $e$ is computed as $l/(1+\alpha \cdot p)^\beta$ and the value of $\rho$ at epoch $e$ is computed as $2/(1+e^{-\gamma \cdot p}) - 1$ where $\alpha$, $\beta$ and $\gamma$ are hyperparameters.

For the BERT baseline we fine-tune for $10$ epochs using the implementation of BERT from the Hugging Face Transformers library\footnote{\url{https://huggingface.co/transformers/}}. We use a batch size of $16$ and a learning rate of $2e-5$ with linear decay after the first $10\%$ of training steps. We optimize using AdamW. To prevent exploding gradients, we apply gradient clipping to $1.0$.

We report validation performance of our models on stance classification (see Table~\ref{tab:devresults}) as well as the score of the topic-discriminator on the training set, since it is not computed on the development set (see Table~\ref{tab:advscores}). We also show the average number of parameters and runtime for all models averaged over all topics (see Table~\ref{tab:time}).

\begin{table*}[t]
    \centering
    \begin{tabular}{l|cccccc}
        \hline
        & DT & HC & FM & LA & A & CC\\ \hline
        BERT & 45.0 & 42.8 & 41.8 & 42.9 & 42.9 & 41.5 \\
        BiCond & \begin{tabular}[t]{@{}l@{}} 66.7\\ \small{(64.6 $\pm$ 0.01)}\end{tabular} & \begin{tabular}[t]{@{}l@{}} 68.7\\ \small{(65.8 $\pm$ 0.01)}\end{tabular} & \begin{tabular}[t]{@{}l@{}} 65.6\\ \small{(64.2 $\pm$ 0.006)}\end{tabular} & \begin{tabular}[t]{@{}l@{}}66.9\\ \small{(64.4 $\pm$ 0.02)}\end{tabular} & \begin{tabular}[t]{@{}l@{}}64.8\\ \small{(62.0 $\pm$ 0.01)}\end{tabular} & \begin{tabular}[t]{@{}l@{}}61.3\\ \small{(59.7 $\pm$ 0.006)}\end{tabular} \\ 
        TOAD &  \begin{tabular}[t]{@{}l@{}} 66.1 \\ \small{(64.4 $\pm$ 0.09)} \end{tabular} & \begin{tabular}[t]{@{}l@{}} 65.9 \\\small{(64.1 $\pm$ 0.18)}\end{tabular} & \begin{tabular}[t]{@{}l@{}} 62.2\\ \small{(59.8 $\pm$ 0.09)}\end{tabular} & \begin{tabular}[t]{@{}l@{}}64.9\\ \small{(63.0 $\pm$ 0.07)}\end{tabular} & \begin{tabular}[t]{@{}l@{}}64.6\\ \small{(62.0 $\pm$ 0.06)}\end{tabular} & \begin{tabular}[t]{@{}l@{}}64.8\\ \small{(58.8 $\pm$ 0.13)}\end{tabular}\\
        $-$ adv & \begin{tabular}[t]{@{}l@{}}69.3\\ \small{(68.1 $\pm$ 0.008)} \end{tabular} & \begin{tabular}[t]{@{}l@{}} 72.6 \\ \small{(70.7 $\pm$ 0.008)}\end{tabular} & \begin{tabular}[t]{@{}l@{}} 68.2\\ \small{(66.7 $\pm$ 0.007)} \end{tabular} & \begin{tabular}[t]{@{}l@{}} 69.2\\ \small{(66.8 $\pm$ 0.01)}\end{tabular} & \begin{tabular}[t]{@{}l@{}}66.5\\ \small{(65.3 $\pm$ 0.006)}\end{tabular} & \begin{tabular}[t]{@{}l@{}} 65.3\\ \small{(63.9 $\pm$ 0.02)}\end{tabular} \\
        \hline
    \end{tabular}
    \caption{$F_{avg}$ results on the development sets for each topic, with mean and variance shown for models with hyperparameter tuning.}
    \label{tab:devresults}
\end{table*}
\begin{table}[t]
    \centering
    \begin{tabular}{l|cccccc}
    \hline
         & DT & HC & FM & LA & A & CC \\
         \hline
        TOAD & 1.9 & 2.2 & 28.7 & 1.3 & 26.5 & 4.2\\ \hline
    \end{tabular}
    \caption{Topic-discriminator F1 on the training set for TOAD across topics.}
    \label{tab:advscores}
\end{table}
\begin{table*}[t]
    \centering
    \begin{tabular}{l|rrrr}
        \hline
         & BERT & BiCond & TOAD & TOAD $-$adv \\
         \hline
        \# parameters & 110 million & 358926 & 554152 & 632915\\
        avg. runtime & 15min & 5min & 20min & 20min\\
        \hline
    \end{tabular}
    \caption{Search trials, time, and parameters for models. We average across all six topics for each model.}
    \label{tab:time}
\end{table*}

\subsection{Hyperparameters}
\label{sec:apphyper}
We tune the hyperparameters for our adversarial model using uniform sampling on the development set with $20$ search trials. We select the best hyperparameter setting using the average rank of the stance classifier F1 (higher is better) and topic discriminator F1 (lower is better). We remove settings where the discriminator F1 is $<0.01$, under the assumption that such low performance is the result of overly corrupt representations that will not generalize. In all models, we use pre-trained 100-dimensional GloVe vectors~\citep{pennington2014glove} in our models. We show hyperparameter configurations and search space for TOAD (Table~\ref{tab:toadhyps}), TOAD without the adversary (Table~\ref{tab:noadvhyps}) and BiCond (Table~\ref{tab:bicondhyps}). Note that there are no hyperparemters to tune for the BERT baseline. 

\begin{table*}[t]
    \centering
    \scalebox{0.9}{
    \begin{tabular}{lc|rrrrrr}
        \hline
        Hyperparameter & Search space &  \multicolumn{6}{c}{Best assignment}\\
        & & DT & HC & FM & LA & A & CC\\
        \hline
        BiLSTM hidden size & \textit{unifrom-integer}[40-150] & 80 & 105 & 113 & 115 & 111 & 105\\
        Stance classifier hidden size  & \textit{uniform-integer}[80-300] & 147 & 278 & 201 & 222 & 213 & 254\\
        Topic discriminator hidden size & \textit{uniform-integer}[40-150] & 85 & 95 & 140 & 120 & 143 & 90\\
        $\lambda_{rec}$ & \textit{choice}[1] & 1 & 1 & 1 & 1 & 1 & 1\\
        $\lambda_{tr}$ & \textit{choice}[0.1, 1, 10] & 0.1 & 0.1 & 10.0 & 10.0 & 1.0 & 0.1\\
        $\gamma$ & \textit{uniform-integer}[10-15] & 14 & 12 & 14 & 11 & 11 & 10\\
        $\alpha$ & \textit{choice}[10] & 10 & 10 & 10 & 10 & 10 & 10\\
        $\beta$ & \textit{choice}[0.25] & 0.25 & 0.25 & 0.25 & 0.25 & 0.25 & 0.25\\
        $l$ & \textit{choice}[0.001] & 0.001 & 0.001 & 0.001 & 0.001 & 0.001 & 0.001\\
        \hline
    \end{tabular}
    }
    \caption{Hyperparameter search space and best settings for TOAD.}
    \label{tab:toadhyps}
\end{table*}
\begin{table*}[t]
    \centering
    \scalebox{0.90}{
    \begin{tabular}{lc|rrrrrr}
        \hline
        Hyperparameter & Search space &  \multicolumn{6}{c}{Best assignment}\\
        & & DT & HC & FM & LA & A & CC\\
        \hline
        BiLSTM hidden size & \textit{unifrom-integer}[40-150] & 96 & 96 & 140 & 134 & 140 & 115 \\
        Stance classifier hidden size  & \textit{uniform-integer}[80-300] & 137 & 137 & 228 & 166 & 228 & 222 \\
        \hline
    \end{tabular}}
    \caption{Hyperparameter search space and best settings for TOAD without the adversary.}
    \label{tab:noadvhyps}
\end{table*}
\begin{table*}[t]
    \centering
    \scalebox{0.9}{
    \begin{tabular}{lc|rrrrrr}
        \hline
        Hyperparameter & Search space &  \multicolumn{6}{c}{Best assignment}\\
        & & DT & HC & FM & LA & A & CC\\
        \hline
        BiLSTM hidden size & \textit{unifrom-integer}[40-150] & 74 & 94 & 128 & 78 & 141 & 104 \\
        Dropout  & \textit{uniform-float}[0.1-0.4] & 0.2380 & 0.3220 & 0.4015 & 0.3086 &  0.3912 & 0.2501 \\
        \hline
    \end{tabular}}
    \caption{Hyperparameter search space and best setting for BiCond.}
    \label{tab:bicondhyps}
\end{table*}
\begin{table}[ht]
    \centering
    \begin{tabular}{lllll}
        \hline
        Topic & \%Pro & \%Con & \%Neither & \# Total \\
        \hline
        DT & 20.9 & 42.3 & 36.8 & 707\\
        HC & 16.6 & 57.4 & 26.0 & 984\\
        FM & 28.2 & 53.8 & 18.0 & 949\\
        LA & 17.9 & 58.3 & 23.8 & 933\\
        A & 16.9 & 63.3 & 19.8 & 733\\ 
        CC & 59.4 & 4.6 & 36.0 & 564 \\
        \hline
    \end{tabular}
    \caption{Class distributions for each of the six topics.}
    \label{app:classdist}
\end{table}

\subsection{Data}
We preprocess tweets by removing URLs and mentions. We remove the $\#$ symbol from hashtags in tweets and tokenize the hashtags. We remove emojis and punctuation from tweets. We convert tweets to lowercase and remove stopwords from the tweets. We show the class distribution in Table~\ref{app:classdist}.

\subsection{Topic Divergence}
\label{app:topicdiv}
Jensen-Shannon divergence~\citep{Lin1991DivergenceMB} is a smoothed, symmetric variant of KL divergence.
Let $t^{(1)}$ and $t^{(2)}$ be two topics and $P$ and $Q$ be word-distributions for the topics respectively. Then the KL divergence is defined as $D_{KL}(P || Q) = \sum_{i} p_i \log \frac{p_i}{q_i}$. However, $D_{KL}(P || Q)$ is undefined if $q_i=0$ for any $q_i \in Q$.  Therefore, Jensen-Shannon divergence uses the average distribution $M = \frac{1}{2}(P + Q)$ and is defined as $$D_{JS}(P || Q) = \frac{1}{2}(D_{KL}(P || M) + D_{KL}(Q || M)).$$

We follow~\citet{Plank2011EffectiveMO} in computing word distributions for each topic pair. Let $V = \cup_t V_t$ be the union of the vocabularies for all topics $t$. Then for the topic pair $t^{(1)}$ and $t^{(2)}$, the distribution for one topic is either $t \in \mathbb{R}^{|V_{t^{(1)}} \cup V_{t^{(2)}}|}$  or $t \in \mathbb{R}^{|V_{t^{(1)}}|}$, where $t_i$ is the probability of the $i$-th word in the vocabulary. 
Note, we use $V_{t^{(1)}} \cup V_{t^{(2)}}$ or $V_{t^{(1)}}$ rather than $V$ to ensure that $m_i \neq 0$ for all $m_i \in M$, regardless of choice of topics. Also note that when using only the vocabulary from $t^{(1)}$, $D_{JS}(P || Q)$ is no longer symmetric, since the size of $t$ depends of which topic is $t^{(1)}$. 

\subsection{Ablation Results}
We report full ablation results on all components of the adversarial model, on all six topics on the development sets (see Table~\ref{tab:fullablation}). 

We also report the results of applying K-means clustering on the representations extracted from the models trained in each setup. For clustering, we extract representations for the entire dataset (train, dev, and test). Then we randomly split the dataset into train and test with no zero-shot topic. We fit K-means clustering on the training portion and evaluate on the test portion. We use the same train/test split for all clusterings. We evaluate using homogeneity (evaluates whether each cluster contains only examples of one topic) and completeness (all examples from one topic are in one cluster) (see Table~\ref{tab:fullclustering}).

\begin{table*}[t]
    \centering
    \scalebox{0.9}{
    \begin{tabular}{l|cc|cc|cc|cc|cc|cc}
    \hline
        & \multicolumn{2}{c|}{DT} & \multicolumn{2}{c|}{HC} & \multicolumn{2}{c|}{FM} & \multicolumn{2}{c|}{LA} & \multicolumn{2}{c|}{A} & \multicolumn{2}{c}{CC} \\
        & F1 & $\Delta$ & F1 & $\Delta$ & F1 & $\Delta$ & F1 & $\Delta$ & F1 & $\Delta$ & F1 & $\Delta$\\ \hline
        TOAD & 49.5 & & 51.2 & & 54.1 & & 46.2 & & 46.1 & & 30.9 & \\ \hdashline
        $-$ transformation & 38.8 & -10.7 & 43.2 & -8.0 & 51.9 & -2.2 &  43.6 & -2.6 & 44.5 & -1.6 & 44.4 & +13.5 \\
        $- \mathcal{L}^{tr}$ & 34.8 & -14.7 & 48.0 & -3.1 & 47.9 & -6.2 & 39.7 & -6.4 & 41.9 & -4.3 & 4.5 & -26.4\\
        \hdashline
        $- \mathcal{L}_t^{rec}$ & 44.6 & -4.9 & 52.5 & +1.3 & 35.0 & -19.1 & 46.9 & +0.7 & 38.7 & -7.4 & 4.4 & -26.5\\
        $- \mathcal{L}_d^{rec}$ & 36.9 & -12.6 & 46.3 & -4.9 & 49.7 & -4.4 & 48.3 & +2.1 & 43.1 & -3 & 18.1 & -12.8\\
        $- \mathcal{L}_t^{rec}$ \& $- \mathcal{L}_d^{rec}$ & 43.0 & -6.5 & 43.5 & -7.7 & 40.1 & -14.0 & 43.3 & -2.9 & 39.5 & -6.6 & 37.3 & +6.4 \\
        \hdashline
        $-$ residual topic & 39.3 & -10.2 & 43.4 & -7.8 & 45.4 & -8.7 & 43.3 & -2.9 & 44.6 & -1.5 & 37.3 & +6.4\\
        $- D^{ul}$ & 40.0 & -9.5 & 51.1 & -0.1 & 44.0 & -10.1 & 46.2 & -0.0 & 40.3 & -5.8 & 26.1 & -4.8\\
        \hline
    \end{tabular}
    }
    \caption{Full component ablation on test sets for all six topics.}
    \label{tab:fullablation}
\end{table*}
\begin{table*}[t]
    \centering
    \scalebox{0.9}{
    \begin{tabular}{l|cc|cc|cc|cc|cc|cc}
        \hline
        & \multicolumn{2}{c|}{DT} & \multicolumn{2}{c|}{HC} & \multicolumn{2}{c|}{FM} & \multicolumn{2}{c|}{LA} & \multicolumn{2}{c|}{A} & \multicolumn{2}{c}{CC} \\
         & Hom. & Com. & Hom. & Com. & Hom. & Com. & Hom. & Com. & Hom. & Com. & Hom. & Com. \\
         \hline
        TOAD & 0.034 & 0.034 & 0.118 & 0.120 & 0.293 & 0.302 & 0.091 & 0.093 & 0.091 & 0.092 & 0.144 & 0.149 \\
        $-$ adv & 0.102 & 0.104 & 0.135 & 0.142 & 0.078 & 0.081 & 0.075 & 0.078 & 0.033 & 0.034 & 0.097 & 0.1\\
        \hline
    \end{tabular}}
    \caption{Homogeneity (Hom.) and completeness (Com.) for clusters computed with the representations extracted from models with each of the six topics as the test set.}
    \label{tab:fullclustering}
\end{table*}

\end{document}